\documentclass[10pt,journal,compsoc]{IEEEtran}
\usepackage{times}
\usepackage{epsfig}
\usepackage{graphicx}
\usepackage{amsmath}
\usepackage{amssymb}
\usepackage{multirow}
\usepackage{colortbl}
\usepackage{listings}
\usepackage{footnote}
\usepackage{algorithm}
\usepackage{algpseudocode}
\usepackage{bm}
\usepackage{cite}
\usepackage{booktabs}
\usepackage[table,xcdraw]{xcolor}
\usepackage{hyperref}

\begin{document}
\title{CANeRV: Content Adaptive Neural Representation for Video Compression}

\author{Lv Tang~\IEEEmembership{Student Member,~IEEE}, Jun Zhu~\IEEEmembership{Student Member,~IEEE},
Xinfeng Zhang~\IEEEmembership{Senior Member,~IEEE}, Li Zhang~\IEEEmembership{Senior Member,~IEEE}, Siwei Ma~\IEEEmembership{Fellow,~IEEE} and Qingming Huang~\IEEEmembership{Fellow,~IEEE}
\thanks{Lv Tang, Jun Zhu, Xinfeng Zhang and Qingming Huang are with School of Computer Science and Technology, University of Chinese Academy of Sciences, Beijing, China. Li Zhang is with Bytedance Inc., San Diego, USA. Siwei Ma is with School of Computer Science, Peking University, Beijing, China.
(Email: luckybird1994@gmail.com, zhujun23@mails.ucas.ac.cn, xfzhang@ucas.ac.cn, lizhang.idm@bytedance.com, swm@pku.edu.cn and qmhuang@ucas.ac.cn.)}}


\IEEEtitleabstractindextext{
\begin{abstract}
Recent advances in video compression introduce implicit neural representation (INR) based methods, which effectively capture global dependencies and characteristics of entire video sequences. Unlike traditional and deep learning based approaches, INR-based methods optimize network parameters from a global perspective, resulting in superior compression potential. However, most current INR methods utilize a fixed and uniform network architecture across all frames, limiting their adaptability to dynamic variations within and between video sequences. This often leads to suboptimal compression outcomes as these methods struggle to capture the distinct nuances and transitions in video content. To overcome these challenges, we propose Content Adaptive Neural Representation for Video Compression (CANeRV), an innovative INR-based video compression network that  adaptively conducts structure optimisation based on the specific content of each video sequence. To better capture dynamic information across video sequences, we propose a dynamic sequence-level adjustment (DSA). Furthermore, to enhance the capture of dynamics between frames within a sequence, we implement a dynamic frame-level adjustment (DFA). {Finally, to effectively capture spatial structural information within video frames, thereby enhancing the detail restoration capabilities of CANeRV, we devise a structure level hierarchical structural adaptation (HSA).} Experimental results demonstrate that CANeRV can outperform both H.266/VVC and state-of-the-art INR-based video compression techniques across diverse video datasets.
\end{abstract}
}

\maketitle
\IEEEdisplaynontitleabstractindextext
\IEEEpeerreviewmaketitle

\section{Introduction}
\IEEEPARstart{W}{ith} the development of digital media, videos have become a fundamental component of modern communication systems, such as short videos, surveillance videos and conference video recordings. As a result, efficiently storing and transmitting videos poses a significant challenge due to the exponential growth of video amount. To address this, a variety of video compression standards have been developed, rooted in classical hybrid video coding frameworks. Notable examples include H.264/AVC~\cite{DBLP:journals/tcsv/WiegandSBL03}, H.265/HEVC~\cite{DBLP:journals/tcsv/SullivanOHW12}, and H.266/VVC~\cite{bross2021overview}. Additionally, the advent of deep learning also has catalyzed significant innovations in video compression techniques, as highlighted by several seminal deep learning based video compression works~\cite{DBLP:conf/dcc/CuiZZJZWZ17,DBLP:journals/pami/LuZO0G021,DBLP:journals/tcsv/YanLLLLW19,DBLP:conf/cvpr/LuO0ZCG19,DBLP:conf/cvpr/Lin0L020,DBLP:conf/cvpr/HuL021,li2021deep,ho2022canf,Li_2024_CVPR,DBLP:journals/tmm/ShengLLLLL23,DBLP:conf/cvpr/LiLL23,DBLP:journals/pami/HuXLJWL23}. These advances are driving the continuous evolution of the video compression task.

\begin{figure}[!t]
    \centering
    \includegraphics[width=\linewidth]{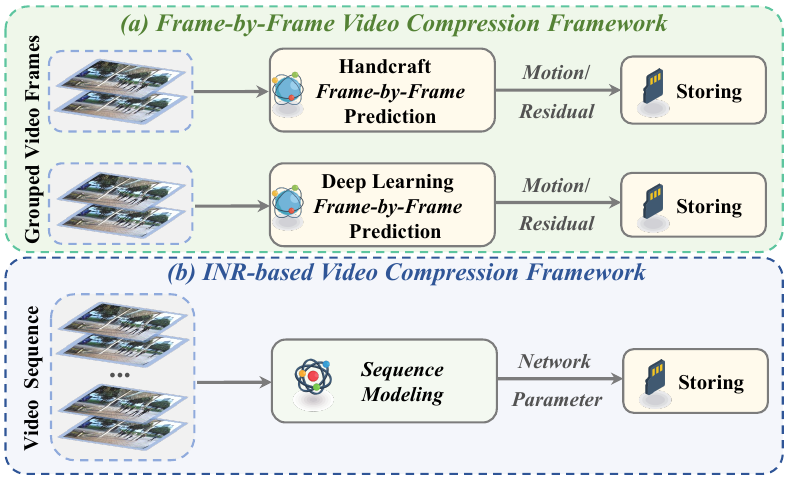}
    \caption{The architecture of existing frame-by-frame style and INR-Based methods. Frame-by-frame style methods may typically contain hybrid video coding methods and deep learning based video coding methods.}
    \vspace{-0.7cm}
    \label{introd1}
\end{figure}

In video compression, both traditional hybrid video coding frameworks and deep learning based video coding frameworks generally follow a similar structure, as illustrated in Fig. \ref{introd1} (a). These frameworks predominantly utilize a frame-by-frame prediction approach, encoding differences between consecutive frames through inter-prediction techniques such as motion estimation and compensation, complemented by entropy encoding of prediction residuals \cite{zhang2018improved,DBLP:journals/tip/WangWZWM19,DBLP:conf/cvpr/LuO0ZCG19,DBLP:conf/mm/Li0022,DBLP:conf/cvpr/Lin0L020,DBLP:conf/cvpr/HuL021,ho2022canf}. While these methods achieve notable performance, they typically underutilize global correlations across entire video sequences, suggesting potential avenues for video compression performance enhancement.

To overcome the aforementioned limitations, recent studies \cite{DBLP:conf/iccv/Tang0ZM23,DBLP:conf/nips/KwanGZGB23,DBLP:conf/cvpr/ChenGLS23a,DBLP:conf/cvpr/0004Y0WCHRLS23} have proposed a sequence modeling framework, illustrated in Fig. \ref{introd1} (b), which represents a paradigm shift in video compression. This approach eschews traditional motion encoding or inter-frame difference computation in favor of employing advanced Implicit Neural Representation (INR) networks. These networks are designed to capture and represent the global dependencies and characteristics across the entire video sequence. By modeling the video sequence holistically, the sequence modeling framework enhances compression performance significantly, outperforming traditional frame-by-frame prediction methods.

\begin{figure*}[!t]
    \centering
    \includegraphics[width=\linewidth]{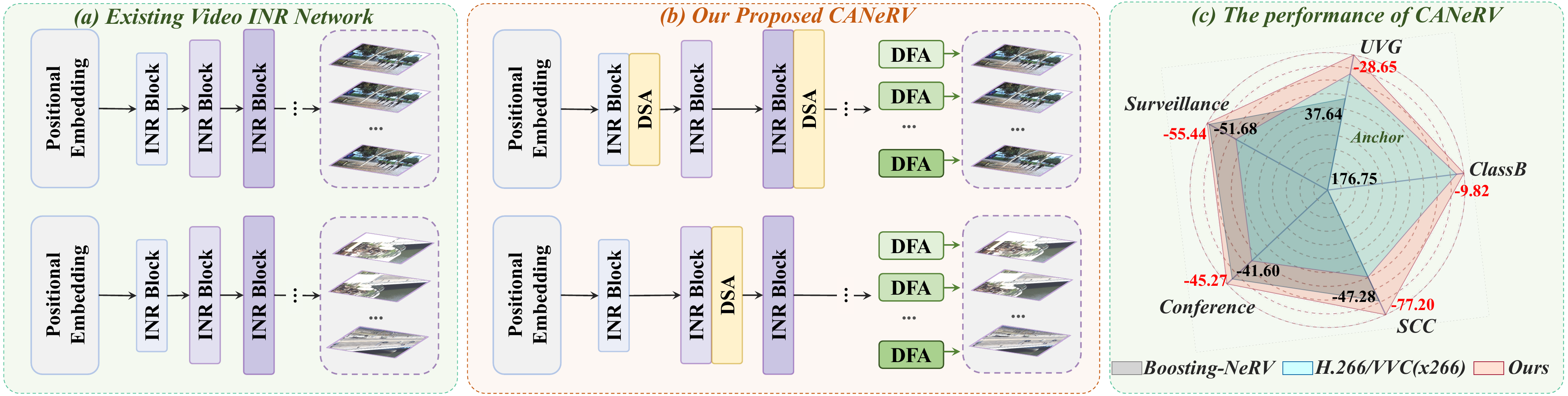}
    \caption{(a) shows that existing INR-based video compression methods use a uniform and fixed architecture configuration to process different videos. (b) is our proposed CANeRV that adaptively optimises the structure of the INR network. (c) is the compression performance of CANeRV.}
    \label{introd2}
    \vspace{-0.5cm}
\end{figure*}

Enhancing the representational capability of networks is crucial for INR-based video compression methods. Current advancements primarily concentrate on refining position encoding techniques and network architectures to effectively represent video sequences. For instance, NeRV \cite{DBLP:conf/nips/ChenHWRLS21} employs a fixed position encoding function across different frames to encapsulate frame-wise temporal information. Expanding on this, HNeRV \cite{DBLP:conf/cvpr/ChenGLS23a} introduces a learnable network that dynamically extracts frame-specific representations, thereby augmenting the model’s adaptability to the video content. Further, HiNeRV \cite{DBLP:conf/nips/KwanGZGB23} enhances this model by incorporating additional temporal information into the position encoding, thereby improving its capability to represent complex temporal dynamics. From an architectural standpoint, E-NeRV \cite{DBLP:conf/eccv/LiWPXML22} incorporates an extra temporal information reconstruction network within the INR framework, and HiNeRV innovates the design of its upsampling modules to boost INR network performance. 
Overall, these approaches significantly elevate the representational efficiency of INR networks, and enhance the overall video compression performance within the INR framework.

{
Despite above advancements, existing INR-based compression methods \cite{DBLP:conf/nips/ChenHWRLS21,DBLP:conf/eccv/LiWPXML22,DBLP:conf/iccv/Tang0ZM23,DBLP:conf/nips/KwanGZGB23,DBLP:conf/cvpr/ChenGLS23a,DBLP:conf/cvpr/0004Y0WCHRLS23} typically rely on a fixed and uniform architecture configuration for different videos, as shown in Fig. \ref{introd2} (a). In particular, the fixed nature of network architecture means the existing INR network architecture is not be tailored to different types of video content, limiting its effectiveness across various video sequences. For example, structures that are effective for sequences with static shots and simple motion might not perform well in scenarios involving complex movements and varied backgrounds. This limitation indicates the necessity for more adaptable INR networks that can dynamically adjust their structures to better adapt to the diversity among video sequences. The uniform architecture configuration employs a consistent set of network parameters across all video frames, severely restricting the INR network ability to adapt to the unique characteristics and dynamic variations of each frame. Such a configuration fails to capture the nuanced transitions between frames, often overlooking subtle changes and finer details, which compromises the overall video quality and fidelity. In addition, the learning process of INR tends to mainly represent the smooth information with low frequency and is difficult to adaptively learn the structure information with relatively high frequency, \textit{e.g.}, the edges within frames. Therefore, addressing above limitations can further enhance the representation ability of INR by improving its adaptability to diverse video content across video sequences, frames, and within frames.
}

Herein, we propose the Content Adaptive Neural Representation for Video Compression (CANeRV), a novel INR network designed to dynamically optimize its structure based on the content characteristics of each video sequence and each video frame, as depicted in Fig. \ref{introd2} (b). CANeRV’s adaptive mechanism aims to adapt to dynamic content across video sequences and frames to improve the representation capabilities of inter-frame differential information. This enables it to handle various types of video content and produce higher-quality reconstructions. Firstly, we design a dynamic sequence-level adjustment (DSA) mechanism that allows for flexible network structure adjustments during the learning process, adapting the INR network structure to content variations in different video sequences. Secondly, we propose a dynamic frame-level adjustment (DFA) mechanism that enhances adaptability between frames by enabling the INR network to learn and adapt to the unique characteristics of each frame, effectively capturing dynamic content variations.
{To further improve the network’s ability to capture spatial structural information within video frames, we propose a structure level hierarchical structural adaptation (HSA) mechanism, which employs an additional learnable network to capture structural information within video frames, enriching detail recovery and enabling the production of higher-quality video frames.} Through our proposed adaptive mechanisms, CANeRV reconstructs higher-quality videos using the similar amount of parameters, thereby improving the rate-distortion (RD) performance of the overall compression framework. As demonstrated in Fig. \ref{introd2} (c), CANeRV significantly outperforms both H.266/VVC and the state-of-the-art INR-based video compression method Boosting-NeRV (CVPR2024) \cite{zhang2024boosting} across diverse video datasets, marking a
significant advancement in the field of video compression. Our main contributions are:

\begin{itemize}

\item {We analyze limitations of INR-based video compression due to the inflexible network architecture, and propose a content adaptive neural representation for video compression (CANeRV). The proposed framework improves INR adaptivity from three levels, \textit{i.e.}, video sequence level, frame level and structure level within frames.}

\item For CANeRV, we propose three key mechanisms: DSA, DFA and HSA. The first two mechanisms improve video compression performance via optimizing the INR network architecture to make it adapt to variations among video sequences and video frames. The last mechanism improves representation capability by enhancing the capture of spatial structural information within frames. 

\item We conduct extensive experiments for our proposed method on video sequences with diverse characteristics including natural videos, surveillance videos, conference videos and screen content videos, and verify the superior performance of the proposed CANeRV. Furthermore, we also analyze the advantages of the INR based video compression on specific video content, which may provide insights for other researchers.

\end{itemize}

\section{Related works}
{
In this section, we first introduce existing frame-by-frame style 
video compression methods, which may typically contain hybrid
video compression methods and deep learning based video compression methods. Then we introduce INR-based image compression methods and video compression methods.}

\begin{figure*}[!t]
    \centering
    \includegraphics[width=\linewidth]{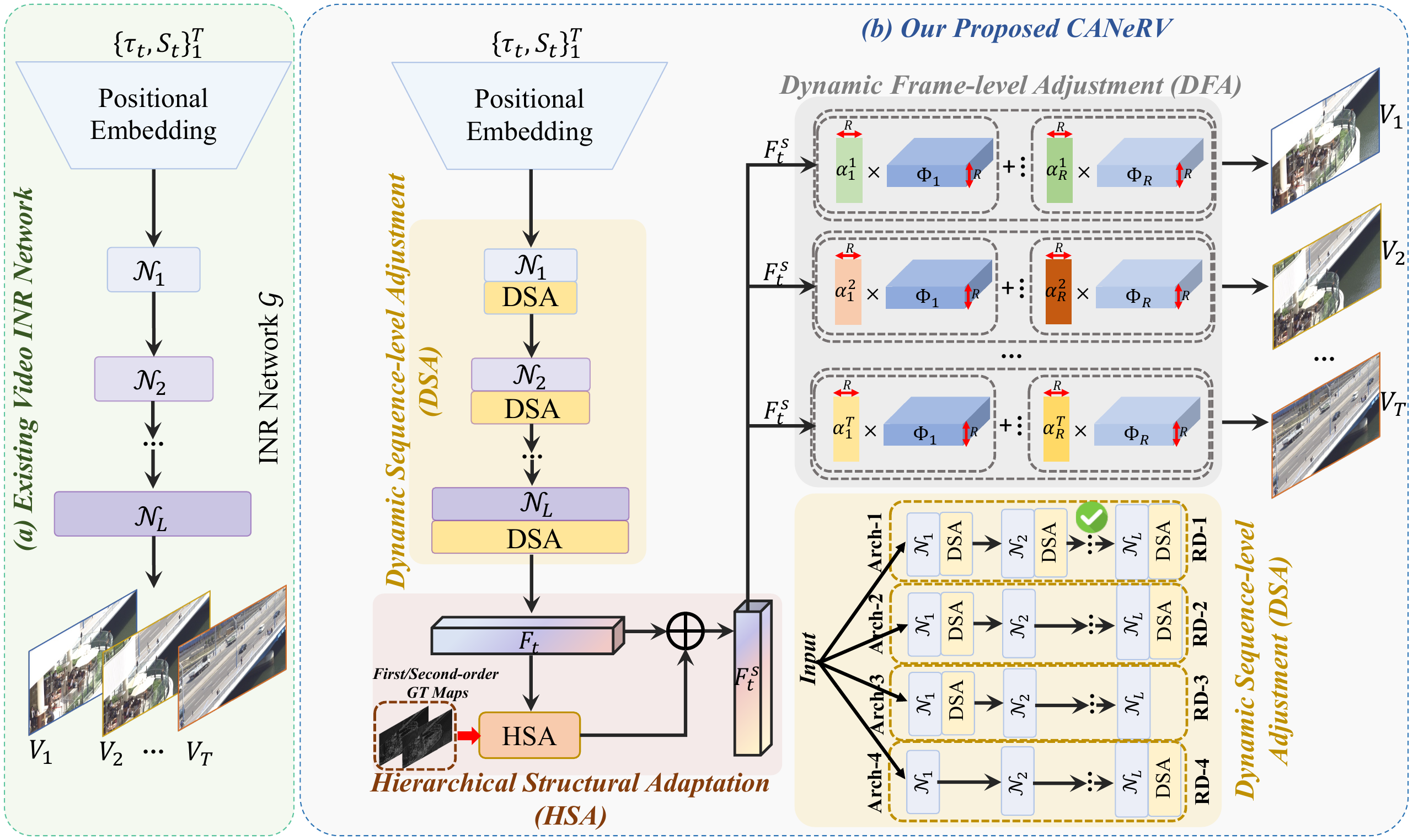}
    \caption{(a) shows the typical architecture of existing video INR network. (b) is the architecture of our proposed novel CANeRV. {For DSA, we briefly hypothesise four architecture adjustment configurations in this figure, with each adjustment yielding the RD performance of the current network architecture. Finally, we select the network architecture that offers the best RD performance.}}
    \label{framework}
\end{figure*}

\subsection{Video Compression} 
\subsubsection{Hybrid Video Compression Framework}
In the past decades, classical video coding standards have predominantly relied on the hybrid video coding framework~\cite{DBLP:journals/tcsv/WiegandSBL03,DBLP:journals/tcsv/SullivanOHW12,bross2021overview}, combining modules such as prediction, transform, quantization, and entropy coding to effectively compress video data. The evolution in this domain has been driven by the enhancement and refinement of lots of coding techniques. These techniques include efficient intra/inter-frame prediction~\cite{cao2012short,DBLP:conf/icip/De-Luxan-Hernandez19,DBLP:journals/tcsv/GaoCECS21,zhang2018improved,DBLP:conf/icmcs/Fu00ZWM020,DBLP:journals/tip/ZhangCZCK19}, multi-core transforms~\cite{zhao2018joint}, implicit transforms~\cite{DBLP:conf/icmcs/ZhangZZL0WM020}, Trellis-Coded Quantization (TCQ)~\cite{DBLP:conf/dcc/SchwarzNMW19}, and loop filtering~\cite{DBLP:journals/jstsp/TsaiCYCHFIWCKL13,zhang2016low}. Furthermore, with the rise of modern video resolutions, the cost of transmitting motion vector information has also increased. To address this, enhanced motion vector coding techniques, such as adaptive precision motion vector coding~\cite{DBLP:conf/pcs/Liu00XWLH19}, history-based motion vector coding~\cite{DBLP:conf/icmcs/YinXZ0LF20}, and decoder-side motion vector refinement~\cite{DBLP:journals/tcsv/GaoCECS21}, have been incorporated into the video coding standards.

Among the techniques discussed, inter-frame prediction is central to video compression, largely determining the efficiency of a video coding framework. This process addresses temporal redundancy by predicting the current frame using the reconstructed ones as reference frames and motion information. In traditional video coding frameworks, inter-frame prediction is primarily achieved through block-level motion estimation (ME) and motion compensation (MC). ME identifies the location in the reference frame that most closely matches the content within the block to be coded, while MC retrieves the content from this location to predict the coding block. Numerous enhancements to block-level ME and MC have been proposed, including the use of multiple reference frames, bi-directional inter prediction (utilizing two reference frames simultaneously), and fractional-pixel ME and MC. Despite these advancements, traditional hybrid video coding frameworks rely on block-level inter-frame prediction, which limits the ability to analyze motion relationships across all frames in a video sequence from a global perspective, thereby impeding optimal inter-frame prediction efficiency. Consequently, many researchers are investigating new coding frameworks that aim to overcome the performance limitations inherent in traditional hybrid video coding frameworks, as discussed in the paper~\cite{DBLP:journals/tcsv/ZhangZFMCK20}.

\subsubsection{Deep-learning based Video Compression Framework}
In contrast to these traditional mechanisms, the advent and rapid maturation of deep learning has opened new avenues in video coding. The integration of deep learning techniques into this field has catalyzed significant advancements in coding efficiency over recent years. As elucidated by comprehensive review studies~\cite{DBLP:journals/csur/LiuLLLW20,DBLP:journals/tcsv/MaZJZWW20}, the strategies leveraging deep learning techniques for video compression can be categorized into two distinct groups: deep-tool methods and deep-framework methods. Deep-tool methods integrate deep neural networks (DNNs) into the established hybrid video coding framework. Their aim is to enhance, or even replace, specific traditional coding tools with their DNN counterparts. Several studies~\cite{DBLP:conf/vcip/ChenLS0CM17,DBLP:journals/tip/LiuYGCJW16,DBLP:conf/vcip/SongLLW17,DBLP:conf/eccv/LuOXZGS18,DBLP:conf/cvpr/YangXWL18,zhao2019enhanced} exemplify this approach, showing improvements in areas such as intra/inter prediction, probability distribution prediction and in-loop filtering. However, there’s a crucial limitation, since the separate coding structure via block-by-block and frame-by-frame prediction cannot fully harness the potential of DNNs. Therefore, although deep-tool methods offer performance enhancements, they are still bounded by the confines of the existing paradigm, which restricts the full potential of improvements that DNNs could provide.

Recent deep-framework methods~\cite{DBLP:conf/eccv/WuSK18,DBLP:conf/iccv/DjelouahCSS19,DBLP:conf/sips/PessoaATF20,DBLP:conf/dcc/CuiZZJZWZ17,DBLP:journals/tcsv/YanLLLLW19,DBLP:conf/cvpr/LuO0ZCG19,DBLP:conf/cvpr/Lin0L020,DBLP:conf/cvpr/HuL021,li2021deep,ho2022canf,DBLP:journals/tmm/ShengLLLLL23,DBLP:conf/cvpr/LiLL23,Li_2024_CVPR} advocate for the development of end-to-end deep learning based video compression frameworks. In these end-to-end deep learning based video compression methods, optical flow estimation algorithms commonly facilitate inter-frame prediction by estimating motion information between adjacent frames~\cite{DBLP:conf/cvpr/LuO0ZCG19,DBLP:conf/cvpr/HuL021,li2021deep}. Moreover, the work~\cite{DBLP:conf/cvpr/Lin0L020} enhances prediction by utilizing multiple reference frames. Additionally, some studies~\cite{DBLP:journals/tmm/ShengLLLLL23} suggest that extracting multi-scale features between two frames can yield more accurate motion information. Despite these methods propose innovative approaches for inter-frame prediction, they remain constrained to a frame-by-frame structure. This limitation reveals the drawbacks observed in traditional hybrid video coding frameworks, particularly the inability to estimate and optimize motion information across all frames in a video sequence from a global perspective.

To address above limitations, our paper proposes a sequence modeling framework CANeRV, that designs the adaptive INR to capture the global dependencies and characteristics throughout the video sequence. By considering the video as a cohesive unit, this framework significantly improves compression efficiency across the entire temporal domain, surpassing the conventional frame-by-frame deep learning based video compression methods.

\subsection{INR-based Image and Video Compression}
INRs represent a novel approach for parameterizing a broad range of signals, fundamentally portraying an object as a function approximated through neural networks. An early example, DeepSDF~\cite{DBLP:conf/cvpr/ParkFSNL19}, provides a neural network-based representation for 3D shapes~\cite{DBLP:conf/cvpr/MeschederONNG19}. The versatility of INRs has recently inspired extensive research across various domains, such as image compression tasks and video compression tasks.

\noindent \textbf{INR-based Image Compression.}
Within compression technologies, INRs have proven to be particularly effective. 
They encode continuous signals directly as learned functions within neural network parameters, offering a distinct and innovative encoding strategy. For instance, INR-based methods have been successfully applied to image compression~\cite{DBLP:journals/corr/abs-2112-04267,dupont2021coin,DBLP:journals/corr/abs-2201-12904,DBLP:journals/spl/ZhangZT23,DBLP:conf/iccv/LadunePHCL23}, revolutionizing traditional approaches. The compression process begins by tightly fitting the INR network to the target image during the encoding phase. Subsequently, the network parameters are quantized and encoded. At the decoder side, a forward pass through the INR network reconstructs the image by calculating RGB values at each spatial position. INR-based codecs employ a simplified, specifically tailored decoder, in contrast to the complex, general-purpose decoders used in autoencoder-based approaches. For example, the COIN decoder~\cite{dupont2021coin} operates with a 10,000-parameter INR network, achieving performance comparable to JPEG~\cite{DBLP:journals/cacm/Wallace91}. The advent of INR-based image compression marks a significant evolution in the image compression field. 
The more comprehensive exploration of image compression can be found in ~\cite{DBLP:journals/corr/TodericiOHVMBCS15,DBLP:conf/iclr/BalleLS17,DBLP:conf/nips/AgustssonMTCTBG17,DBLP:conf/iclr/BalleMSHJ18,DBLP:conf/iclr/TheisSCH17,DBLP:conf/icml/RippelB17,DBLP:conf/iccv/AgustssonTMTG19,DBLP:conf/cvpr/MentzerATTG18,DBLP:conf/nips/MinnenBT18,DBLP:conf/iclr/LeeCB19,DBLP:conf/icml/BlauM19,DBLP:conf/cvpr/ChengSTK20,DBLP:conf/cvpr/ChengFHGWZZS21,DBLP:journals/tip/ChenLMSCW21,DBLP:journals/tip/WangDTXL21,DBLP:journals/pami/MaLYLW22,DBLP:conf/cvpr/HeZSWQ21,DBLP:conf/mm/XieCC21,Jia_2024_CVPR,Zhang_2024_CVPR,DBLP:conf/iccv/YangM23,DBLP:conf/iccv/Park0K23,DBLP:conf/cvpr/LeeJCP022,DBLP:journals/pami/DuanLMHMZ24,DBLP:journals/pami/HuYML22,DBLP:journals/pami/0005ZGY021}.

\noindent \textbf{INR-based Video Compression.}
Furthermore, INRs have been adapted for video compression~\cite{DBLP:journals/corr/abs-2112-11312,DBLP:conf/iccv/Tang0ZM23,DBLP:conf/nips/KwanGZGB23,DBLP:conf/cvpr/ChenGLS23a,DBLP:conf/cvpr/0004Y0WCHRLS23,DBLP:conf/nips/ChenHWRLS21,DBLP:conf/eccv/LiWPXML22}, demonstrating their potential to enhance compression techniques by leveraging the global correlations across video sequences, thus optimizing compression performance more effectively than traditional methods.
In the realm of INR-based video compression, various advancements have been made to enhance the efficiency of these methods. For example, numerous studies have been conducted to improve the representational power of video INR networks through different strategies, such as the patch-wise modeling strategy~\cite{DBLP:conf/cvpr/MaiyaGE0LPWWS23,DBLP:conf/icip/BaiDWY23}. In addition, there is targeted research that focuses on modeling residuals on a volume-wise and frame-wise basis~\cite{DBLP:conf/cvpr/MaiyaGE0LPWWS23,DBLP:conf/cvpr/ZhaoAM23}, as well as incorporating flow-based motion compensation~\cite{DBLP:journals/corr/abs-2112-11312,DBLP:conf/mm/LeeRKP23}. These techniques contribute to scalable encoding and enhance the ability to handle longer and more varied video sequences. Moreover, sophisticated loss functions for INR networks have been explored to improve their representation capability~\cite{DBLP:conf/iccv/Tang0ZM23}. These advancements significantly improve the performance of INR-based video compression technologies.

Despite significant progress in INR-based video compression, current methods generally use a uniform and fixed architecture configuration. This configuration tends to restrict flexibility and efficiency when dealing with varied video content. Addressing these limitations, we propose CANeRV, which can adaptively adjust the INR network structure tailored to the specific content characteristics of each video sequence. CANeRV’s adaptive mechanism allows for the reconstruction of higher-quality videos using the similar amount of parameters as conventional methods. By dynamically adjusting the network to better align with the unique attributes of each video, CANeRV significantly enhances the RD performance of the INR-based video compression method.

\section{Analysis of Video INR}
Herein, we introduce existing INR networks in detail and discuss the limitations of these INR-based video compression methods. This will help readers better understand the insights in this paper.

\subsection{Preliminaries}
As shown in Fig. \ref{framework} (a), existing video INR networks, like works~\cite{DBLP:conf/nips/ChenHWRLS21,DBLP:conf/eccv/LiWPXML22,DBLP:conf/cvpr/ChenGLS23a,DBLP:conf/nips/KwanGZGB23}, employ a neural network $\mathcal{G}$ to reconstruct video frames. These INR-based methods, which utilizes MLPs or CNNs, aim to represent video frames based on a temporal index $t$ for temporal mapping and a normalized grid $S$ for spatial mapping. With continuous development, various technologies have been introduced that further enhance the ability of these INR networks to better model different video sequences, thereby advancing the development of this field.

To simplify the above modeling process, we can explain it with a more intuitive formulation:
\begin{equation}
    V_t = \mathcal{G}(\tau_t,S_t).
\end{equation}
Here, $\tau_t$ and $S_t$ represent the temporal and spatial information of the frame $V_t$, respectively. The reconstruction process via the network layers can be further defined as:
\begin{equation}
    V_t = \mathcal{N}_L(\dots(\mathcal{N}_2(\mathcal{N}_1(\tau_t,S_t)))).
\end{equation}
$\{\mathcal{N}_1, \mathcal{N}_2, \dots, \mathcal{N}_L\}$ denotes the individual reconstruction layer within the network $\mathcal{G}$, and $L$ is the total number of reconstruction layers. For the $l^{th}$ reconstruction layer, it can be defined as:
\begin{equation}
\mathcal{N}_l(x_l^t) = W_l \otimes x_l^t,
\label{ori_layer}
\end{equation}
where $\otimes$ means matrix multiplication and $W_l$ means the trainable weight matrix for the $l^{th}$ reconstruction layer. $x_l$ means the input of $l^{th}$ reconstruction layer, derived from the output of $(l-1)^{th}$ reconstruction layer. Note that, the first reconstruction layer is defined as $x_1^t=(\tau_t,S_t)$.

\subsection{Limitations of Existing methods} 
As shown in Fig. \ref{framework} (a), current video INR networks typically employ a fixed and uniform network $\mathcal{G}$, to model an entire video sequence $\{V_t\}_1^T$, with a set of network parameters shared across all frames. Although this approach may seem efficient, it also has substantial limitations that would restrict its effectiveness. 

{A primary drawback of existing INR networks~\cite{DBLP:conf/cvpr/ChenGLS23a,DBLP:conf/iccv/Tang0ZM23,DBLP:conf/nips/KwanGZGB23} is that they often employ a fixed structure regardless of the video content, leading to inefficiencies, especially when adapting to diverse scenes. For example, a network structure optimized for videos with complex motion and frequent camera transitions may not be optimized for scenes with simpler motion and smoother transitions. Therefore, this fixed design can result in suboptimal RD performance for these networks. Another shortcoming of these networks is their inability to capture the dynamic variations among video frames effectively. Since they rely on a uniform parameter for the entire sequence, subtle changes and movements within the video cannot be adequately represented. Video content is inherently dynamic, characterized by constantly changing scenes and actions. The uniform parameter configuration lacks the necessary flexibility to adapt to these variations, severely restricting the model’s capacity to accurately represent the evolving dynamics of the video content. In addition, INR methods compress videos via overfitting video signals with as few parameters as possible. This strategy forces these parameters to represent the common and smooth contents, which mainly corresponds to the low-frequency components. However, some structural information with high frequency is also very important to reconstruction quality, especially to human visual system, \textit{e.g.}, the edges with in frames. However, existing INR learning process cannot well adapt to these structural information representation.  
}

Given above limitations, there’s an evident need for more flexible modeling approaches that can better adapt to the varied and dynamic characteristics of video content. In this paper, we propose a dynamic architecture adjustment strategy that adapts to specific differences among frames and videos. This approach could significantly enhance the INR models’ ability to represent video content effectively. Such adaptive approaches would allow the INR network to more effectively capture the complex dynamics within videos, marking a significant evolution from static, fixed-architecture models to dynamic, adaptable ones. 
In addition to the dynamic network architecture adjustment mechanism, we also design an intra-frame spatial structure prediction module. This allows the INR network to predict the structural information within the current video frame. The predicted structural information is then used to guide the INR network learning process to enhance the INR representation ability to structural information.

\section{Method: Our Proposed CANeRV}
We firstly provide an overview for the proposed CANeRV, and then we detail the three modular innovations of CANeRV.

\subsection{Overview of Our Proposed CANeRV}
To address limitations identified in current video INR networks, this paper introduces the content adaptive learning mechanism in the INR network and proposes the CANeRV. Our proposed adaptive mechanism consists of three pivotal components, shown in Fig. \ref{framework} (b). The first component is dynamic sequence-level adjustment (DSA). This component enables the INR network structure to flexibly adapt to the various demands of different video sequences. The second component is dynamic frame-level adjustment (DFA) to adapt to the variations of frame-specific content by introducing frame-level parameters. This strategy enhances the network’s ability to capture and accurately represent the subtle nuances and dynamics inherent in each frame. DSA and DFA jointly enable CANeRV to adaptively optimise the network structure based on the video content from both sequence and frame level. To better capture the structural information of videos, thereby reconstructing higher quality video frames, we propose the hierarchical structural adaptation (HSA) mechanism. By incorporating additional network layers that are tasked with learning both first-order and second-order structural information from video frames, HSA further improves the reconstruction quality of our proposed CANeRV.

\subsection{Dynamic Sequence-level Adjustment (DSA)}
As outlined above, existing INR networks often employ a fixed network structure to model various video sequences. In particular, a common trait of these INR networks is that they use the same network structure for modeling different video sequences. Therefore, they may overlook the significant variations in video content and unique structural characteristics for each sequence, resulting in suboptimal performance due to their inability to adapt to the specific demands of different video sequences. For example, a network designed for fast-paced video sequences, may perform poorly for slow-moving video scenes. 

To address above limitation, we propose DSA that tailors the INR network structure dynamically based on the content characteristics of each specific video sequence. Mathematically, after adding DSA to the INR network, the Eqn. \ref{ori_layer} can be re-written:
\begin{equation}
\mathcal{N}_l(x_l^t) = \phi_l(W_l) \otimes x_l^t,
\label{reconstruction}
\end{equation}
where $\phi_l$ represents the adaptive strategy of the network structure for $l^{th}$ layer. By implementing dynamic adaptive strategy, the INR network can become significantly more efficient to different video content. This flexibility ensures that the network always operates at its optimal configuration for one certain video sequence, potentially leading to better RD performance.

\begin{algorithm}[!t]
\caption{Binary Search for Optimal Layer in CANeRV}
\begin{algorithmic}[1]
\State Define layer depths $\{0, 1, 2, 3, 4\}$
\State Compute $\mathcal{L}_0$ and $\mathcal{L}_2$ with depths 0 and 2
\If {$\mathcal{L}_0 < \mathcal{L}_2$}
    \State Compute $\mathcal{L}_1$ with depth 1
    \If {$\mathcal{L}_0 < \mathcal{L}_1$}
        \State Optimal depth $\leftarrow 0$
    \Else
        \State Optimal depth $\leftarrow 1$
    \EndIf
\Else
    \State Compute $\mathcal{L}_3$ with depth 3
    \If {$\mathcal{L}_2 < \mathcal{L}_3$}
        \State Optimal depth $\leftarrow 2$
    \Else
        \State Compute $\mathcal{L}_4$ with depth 4
        \If {$\mathcal{L}_3 > \mathcal{L}_4$}
            \State Optimal depth $\leftarrow 4$
        \Else
            \State Optimal depth $\leftarrow 3$
        \EndIf
    \EndIf
\EndIf
\State \textbf{return} Optimal depth
\end{algorithmic}
\label{algo_binary}
\end{algorithm}

Herein, we should determine the optimal adaptive strategy $\phi_l$ for a specific sequence to ensure that the INR network structure, as modified by this mechanism, achieves optimal RD performance during video sequence compression. Generally, the selection of $\phi_l$ can be formulated as an optimization problem:
\begin{align}
\phi_l^{*} = \underset{\phi_l}{min} \Big( \lambda\underbrace{\hat{D}( \mathcal{Q}(\sum_{l=1}^L\phi_l)}_{D}) + \underbrace{R(\sum_{l=1}^L\mathcal{P}(\phi_l))}_{R} \Big).
\label{equ:optim}
\end{align}
{In Eqn. \ref{equ:optim}, $\phi_l^{*}$ means the final optimization outcome of the formula. $\mathcal{P}(\cdot)$ is utilized to compute the number of parameters within each layer, while $R(\cdot)$ denotes the number of bits required for encoding network parameters. $\mathcal{Q}$ means using all parameters to reconstruct video frames. $\hat{D}$ represents the distortion value of reconstructed video frames, measured using mean squared error (MSE). $\lambda$ is the Lagrange multiplier. During the implementation of this optimization process, we simplify the solution space to find the optimal $\phi_l$. In particular, suppose we have two sets of adjustment strategies, $\{\phi_l^a\}_1^L$ and $\{\phi_l^b\}_1^L$. After these two strategies change the network structure, the parameters contained within the network are $\Theta_a$ and $\Theta_b$, and the reconstructed video frames are $\{V_t^a\}_1^T$ and $\{V_t^b\}_1^T$, respectively. Given the above information, the metric to evaluate the merits of these two adjustment mechanisms can be:

\begin{align}
\mathcal{L}_a = \lambda D_a + R_a &= \lambda d(\{V_t^a\}_1^T,\{V_t\}_1^T) + R(\Theta_a), \\
\mathcal{L}_b = \lambda D_b + R_b &= \lambda d(\{V_t^b\}_1^T,\{V_t\}_1^T) + R(\Theta_b).
\end{align}
$d( , )$ denotes the distortion between reconstruction and original frames. If $\mathcal{L}_a < \mathcal{L}_b$, we select $\{\phi_l^a\}_1^L$. Otherwise, we opt for $\{\phi_l^b\}_1^L$. Therefore, through above approach, we effectively compare the merits of different adjustment strategies.}

Once the methodology for comparing the relative merits of different $\phi_l$ has been established, the subsequent challenge involves defining the potential search space for $\phi_l$. Inspired by neural architecture search (NAS)~\cite{DBLP:journals/corr/abs-2301-08727}, we further constrain the search space of $\phi_l$ in this paper. In particular, we adopt the strategy of deepening an INR layer as a concrete means of adjusting the network structure. We define the potential depths as $\{0, 1, 2, 3, 4\}$. The simplest approach would be to enumerate all possible depths, but this approach is computationally expensive. Therefore, we employ binary search to optimize the process, as shown in Algo. \ref{algo_binary}. The binary search approach in Algo. \ref{algo_binary} significantly streamlines the selection process for optimal network depth adjustments.

\begin{figure}[!t]
    \centering
    \includegraphics[width=\linewidth]{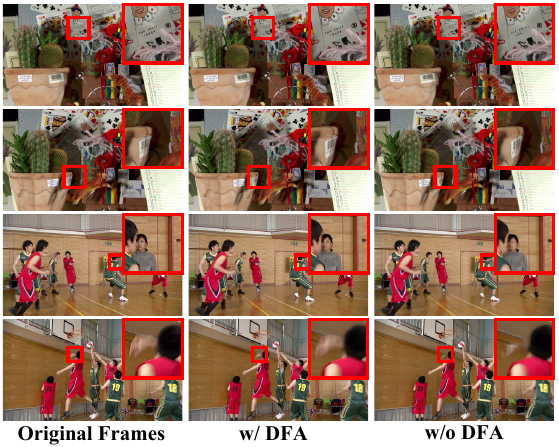}
    \caption{Visual comparison between CANeRV using DFA and not using DFA. w/ means ``with" operation and w/o means ``without" operation. For sequences with complex motion, such as the Basketball sequence, DFA effectively aids CANeRV in capturing the unique characteristics of different frames, thereby reconstructing higher-quality video frames.}
    \label{CFT_Vis}
    \vspace{-0.5cm}
\end{figure}

\subsection{Dynamic Frame-level Adjustment (DFA)}
While the DSA enhances the adaptability of the INR network across various video sequences, it still cannot solve the adaption problem for frame-specific variation. In particular, DSA assumes that a single, albeit dynamically adjusted, network configuration is uniformly applied to all frames of a video sequence. It still depends on shared network parameters to reconstruct each frame. As a result, this method inherently limits the network’s ability to capture the variations specific to individual frames within a video.

To address the above issue and effectively capture content variations unique to each frame, we propose the DFA component, illustrated in Fig. \ref{framework} (b). This innovative feature enhances the existing INR network by adding a specialized network layer designed to independently learn and adapt to the specific characteristics of each individual frame. By enabling frame-specific adaptation, DFA plays a crucial role in accurately capturing the transient dynamics essential for high-quality video reconstruction, effectively overcoming the challenges posed by INR approaches using uniform parameters across varied frames.

Eqn. \ref{reconstruction} gives the mathematical definition of a INR network layer, showing the reconstruction process of each INR layer. 
DFA further advances this reconstruction process by integrating an additional learnable parameter $\Delta W_l^t$ into $W_l$ for the $t^{th}$ frame. Therefore, the revised equation of Eqn. \ref{reconstruction} is written as:
\begin{equation}
\mathcal{N}_l^{\prime}(x_l^t) = (\phi_l(W_l) + \Delta W_l^t) \otimes x_l^t,
\label{dfa_eqn}
\end{equation}
where $\Delta W_l^t$ denotes the additional learnable parameter introduced by DFA, engineered to capture the unique characteristics of the $t^{th}$ frame. The magnitude of $\Delta W_l^t$ is designed to vary depending on the degree of variability between frames. From Fig. \ref{CFT_Vis}, we can see that DFA provides a more detailed and accurate adaptation to different frames within an video sequence.

\noindent \textbf{Parameter Dimensionality Reduction.}
An intuitive approach to realise DFA is to introduce a learnable layer $\Delta W_l^t$ for each video frame, matching the dimensions of $W_l$. However, this approach presents a problem: a video sequence often contains hundreds of frames, thereby necessitating the inclusion of an equal number of $\Delta W_l^t$. This increase in the parameter amount makes the overall network parameters and the learning process uncontrollable and difficult to optimize for an effective INR network. To solve this challenge, we draw inspiration from the concept of low-rank matrix factorization, a method recognized for its effectiveness in reducing parameter amount while preserving essential model capabilities. In particular, this method can be formalized as:
\begin{equation}
\Delta W_l^t = \sum_{r=1}^{R} \alpha_r^t \Phi_r.
\label{low_Rank_Eqn}
\end{equation}
The above formula illustrates how to represent $\Delta W_l^t$ for layer $l$ at time step $t$ as a sum of contributions from multiple low-rank matrices. Each component in this summation consists of a learnable coefficient $\alpha_r^t$ that varies with time, multiplied by a trainable basis matrix $\Phi_r$. The learnable coefficients adjust the influence of each basis matrix, which provides a structured pattern to the weight update. This low-rank approximation method efficiently captures the dynamic changes in the network’s weights by representing the updates within a lower-dimensional subspace formed by these basis matrices. Through low-rank factorization, we can efficiently implement DFA. As illustrated in Fig. \ref{CFT_Vis}, the integration of DFA into our proposed CANeRV enables the reconstruction of higher-quality video frames.

\begin{figure}[!t]
    \centering
    \includegraphics[width=\linewidth]{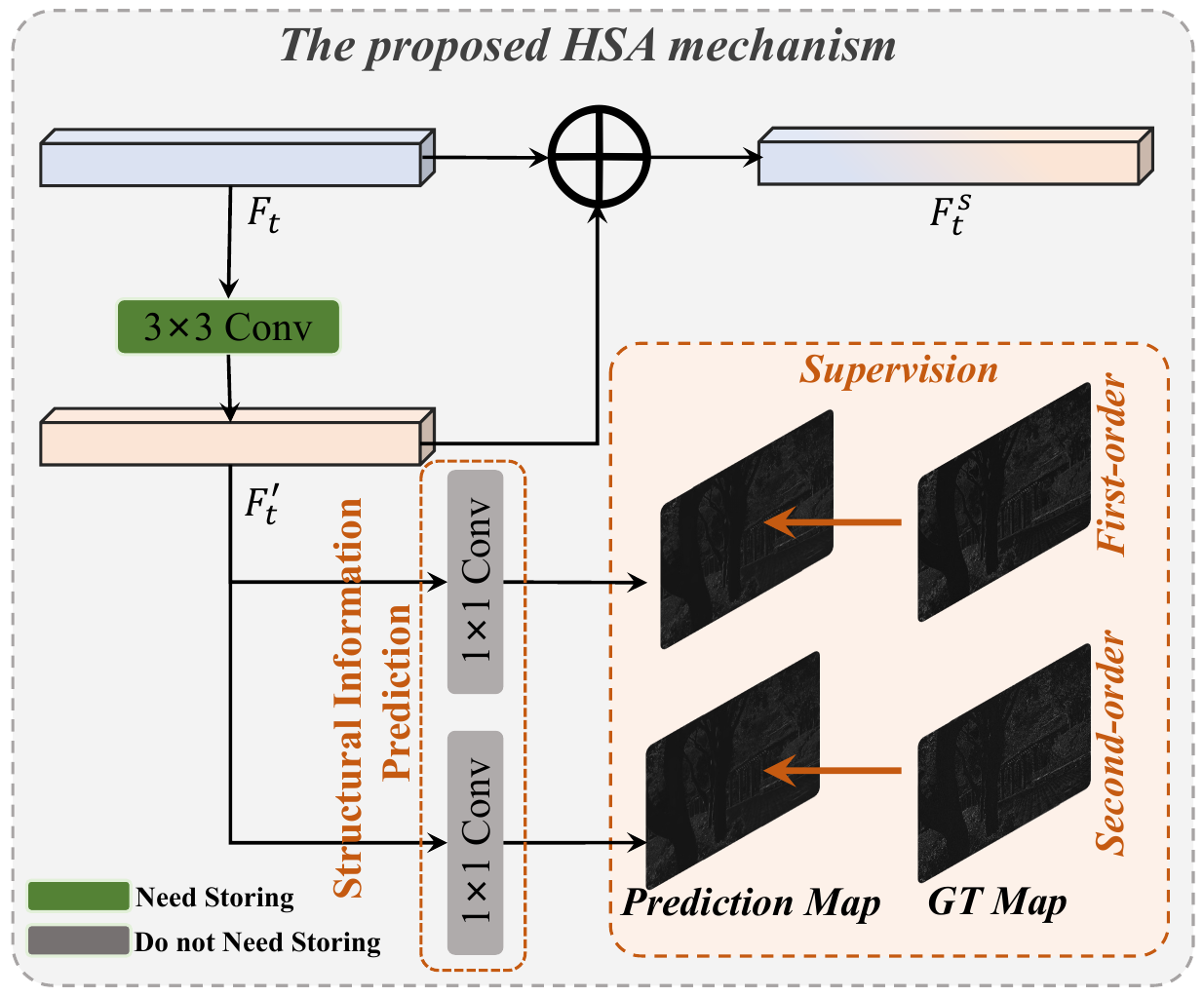}
    \caption{The architecture of our proposed HSA mechanism. In the HSA mechanism, the parameters of the $3\times3$ convolution operations need to be compressed, while the parameters involved in the two $1\times1$ convolution operations do not require to be compressed.}
    \label{HSA_Arch}
    \vspace{-0.5cm}
\end{figure}

\subsection{Hierarchical Structural Adaptation (HSA)}
DSA and DFA, from sequence-level and frame-level perspectives respectively, have been developed to create an adaptive network mechanism that enhances flexibility in adapting to variations in video content. To further improve the capability of our CANeRV to capture the structural details of each video frame, thereby improving the restoration of detailed structural information and enhancing the fidelity of reconstructed videos, we propose the HSA by introducing the constraints on both first-order and second-order structural information of video frames, facilitating the reconstruction of higher-quality video frames.

\begin{figure}[!t]
    \centering
    \includegraphics[width=\linewidth]{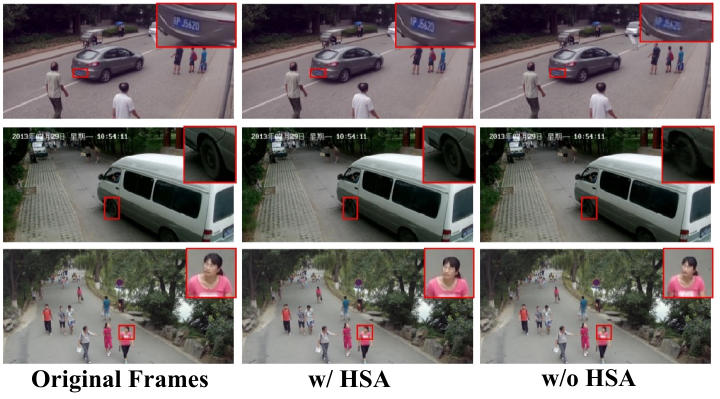}
    \caption{Visual comparison between CANeRV using HSA and not using HSA. The figure demonstrates that HSA further helps CANeRV capture the detailed structural information of each frame, resulting in the reconstruction of higher-quality video frames.}
    \vspace{-0.5cm}
    \label{HSA_Vis}
\end{figure}

The architecture of HSA is shown in Fig. \ref{HSA_Arch}. For the feature $F_t$, originating from the output of the last layer ($L^{th}$ layer) of the INR network, our goal is to capture more detailed structural information from the original video with the minimal increase in parameter amount. To achieve this, we first apply a $3\times3$ convolution to transform the feature $F_t$ to $F'_t$. Then, to ensure that the feature $F'_t$ focus more on reconstructing structural information, we use the current frame’s first-order and second-order structural information for reconstruction. In particular, we employ two $1\times1$ convolutions to predict the first-order structural map and the second-order structural map, supervised by their corresponding ground truths (GTs). For the first-order GT map, we use the Canny operator to extract the first-order structural map from the original video frame, which pertains to gradients and edges within a video frame. For the second-order GT map, we use the Laplacian operator to extract the second-order structural map from the original video frame, capturing the curvature and continuity of edges and textures. Finally, $F’_t$ is added to $F_t$, resulting in the output feature $F_t^s$. Such detailed information capturing is important for dynamic scenes where the texture and edge complexities can significantly influence the perceived quality of the video. Note that parameters of two $1\times1$ convolutions do not need to be compressed. This means that HSA only adds a single $3\times3$ convolution parameter, yet it further promotes the network’s ability to represent structural information. As shown in Fig. \ref{HSA_Vis}, with our proposed HSA, CANeRV can reconstruct better quality video frames that contain more detailed information.

\begin{figure*}[!t]
    \centering
    \includegraphics[width=\linewidth]{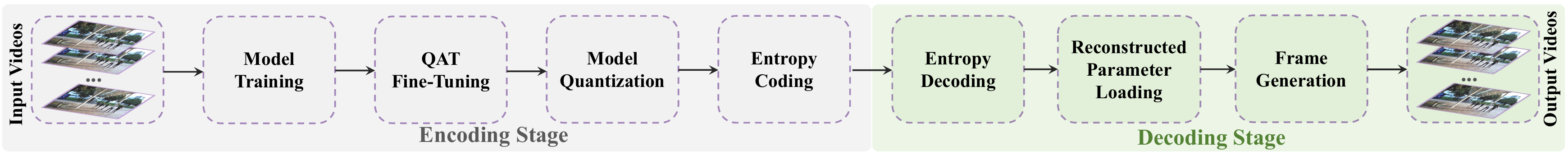}
    \caption{The encoding stage and the decoding stage of our proposed CANeRV.}
    \label{encoding_decoding}
    \vspace{-0.5cm}
\end{figure*}

\begin{table*}[]
\centering
\caption{BDBR(\%) performance of different methods when compared with H.266/VVC (x266) on the common scene video sequences HEVC ClassB and UVG in terms of PSNR and MS-SSIM. Text in {\color[HTML]{FF0000} red font} indicates leading performance compared to other methods, or signifies that as an INR-based method, it has surpassed H.266/VVC (x266) for the first time.}
\begin{tabular}{@{}cccccc@{}}
\toprule
\multicolumn{1}{c|}{Dataset} &
  \begin{tabular}[c]{@{}c@{}}HiNeRV\\ (NeurIPS2023)\end{tabular} &
  \begin{tabular}[c]{@{}c@{}}Boosting-NeRV\\ (CVPR2024)\end{tabular} &
  \multicolumn{1}{c|}{\begin{tabular}[c]{@{}c@{}}Ours\end{tabular}} &
  \begin{tabular}[c]{@{}c@{}}DCVC-DC\\ (CVPR2023)\end{tabular} &
  \begin{tabular}[c]{@{}c@{}}DCVC-FM\\ (CVPR2024)\end{tabular} \\ \midrule
\multicolumn{6}{c}{BDBR(PSNR)}                                                                         \\ \midrule
\multicolumn{1}{c|}{HEVC ClassB} & 4.64   & 176.75 & \multicolumn{1}{c|}{\color[HTML]{FF0000}\textbf{-9.82}}  & -32.05 & -45.01 \\
\multicolumn{1}{c|}{UVG}         & -22.01 & 37.64 & \multicolumn{1}{c|}{\textbf{-28.65}} & -28.99 & -35.99 \\ \midrule
\multicolumn{6}{c}{BDBR(MS-SSIM)}                                                                      \\ \midrule
\multicolumn{1}{c|}{HEVC ClassB} & -25.69      & 110.87 & \multicolumn{1}{c|}{\textbf{-28.19}}               & -24.47      & -38.23      \\
\multicolumn{1}{c|}{UVG}         & -33.17      & 91.10 & \multicolumn{1}{c|}{\color[HTML]{FF0000}\textbf{-40.77}}               & -19.76      & -22.62      \\ \bottomrule
\end{tabular}
\label{bdbr_classb_uvg}
\end{table*}

\begin{figure*}[!t]
    \centering
    \includegraphics[width=\linewidth]{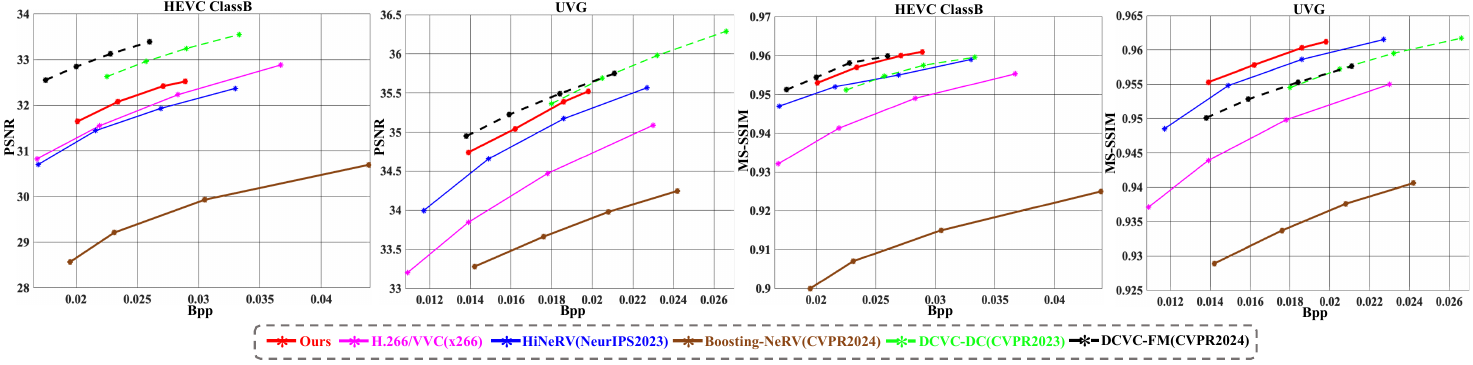}
    \caption{RD curves of our proposed method and other methods on HEVC ClassB and UVG datasets in terms of PSNR and MS-SSIM.}
    \vspace{-0.5cm}
    \label{rd_classb_uvg}
\end{figure*}

\subsection{Implementation Details}
In this paper, we propose three modules, DSA, DFA and HSA. We have selected the following structural design: For the INR network $\mathcal{G}$, we first implement HSA to optimize the feature $F_t$, assisting the INR network in better capturing the detailed structural information of the current video. Then we incorporate DSA to adaptively optimize the INR network structure. 
Finally, with the optimized feature $F_t^s$ and network structure, we incorporate DFA to further capture the unique characteristic of each video frame.

In this paper, we train the CANeRV using the combination of MS-SSIM and MSE losses.
We train our CANeRV for about 300 epochs with Adam optimizer~\cite{DBLP:journals/corr/KingmaB14}. We initialize the learning rate at 5e-4 and use a cosine annealing learning rate schedule~\cite{DBLP:conf/iclr/LoshchilovH17}. After the training, we enhance the CANeRV's performance through quantization-aware-training
(QAT). This approach allows us to quantize the network parameters
to 6 bits without losing significant information. We use QAT to fine-tune the CANeRV for about 30 epochs. After QAT, we quantize the network parameters with 6 bits
and apply arithmetic coding to the quantized network parameters,
transforming them into a compact bitstream suitable for transmission and storage. The above process is the overall encoding stage. The decoding process begins with arithmetic decoding, which retrieves the quantized network parameters. These reconstructed parameters are then reloaded into the CANeRV network. By
performing forward propagation, the INR network reconstructs
the video frames. Note that, in our proposed CANeRV, the design of the INR block is the same as that in HiNeRV. The encoding stage and the decoding stage of our proposed CANeRV is shown in Fig. \ref{encoding_decoding}.

\section{Experiment}
\subsection{Evaluation Databases and Metrics} 
To fully verify the effectiveness of our proposed CANeRV, we conduct the evaluation on different video sequences, which include various real-world scenes. These video sequences primarily include 12 common scenes, of which 5 videos are from HEVC ClassB~\cite{DBLP:journals/tcsv/SullivanOHW12} and 7 videos are from UVG~\cite{DBLP:conf/mmsys/MercatVV20}. Additionally, the video sequences include 3 video conference scenes from HEVC ClassE~\cite{DBLP:journals/tcsv/SullivanOHW12}. Furthermore, we also select 7 surveillance video sequences from works~\cite{ieee1857,DBLP:journals/tcsv/ZhaoWWYMG22}, and 7 screen content coding (SCC) video sequences to further assess the performance of our CANeRV. Through these comprehensive evaluations, the effectiveness of our proposed CANeRV can be fully verified.

In this paper, we use PSNR and MS-SSIM to measure the quality of the reconstructed frames, which are the commonly used quality metrics in video compression. The compression rate is measured by the bits per pixel (Bpp). Additionally, we evaluate various video compression methods using the Bjøntegaard Delta Bit Rate (BDBR)~\cite{bjontegaard2001calculation}. The BDBR is a measure of how much bit rate is saved when compared to the baseline algorithm at the same quality, measured by PSNR and MS-SSIM. 

\begin{table*}[]
\centering
\caption{BDBR(\%) performance of different methods when compared with H.266/VVC (x266) on surveillance, conference and SCC videos in terms of PSNR and MS-SSIM. {\color[HTML]{FF0000} Red fonts} mean achieving leading performace compared to other methods.}
\begin{tabular}{@{}cccccc@{}}
\toprule
\multicolumn{1}{c|}{Dataset} &
  \begin{tabular}[c]{@{}c@{}}HiNeRV\\ (NeurIPS2023)\end{tabular} &
  \begin{tabular}[c]{@{}c@{}}Boosting-NeRV\\ (CVPR2024)\end{tabular} &
  \multicolumn{1}{c|}{\begin{tabular}[c]{@{}c@{}}Ours\end{tabular}} &
  \begin{tabular}[c]{@{}c@{}}DCVC-DC\\ (CVPR2023)\end{tabular} &
  \begin{tabular}[c]{@{}c@{}}DCVC-FM\\ (CVPR2024)\end{tabular} \\ \midrule
\multicolumn{6}{c}{BDBR (PSNR)}                                                                                                 \\ \midrule
\multicolumn{1}{c|}{Surveillance} & -52.68 & -51.68 & \multicolumn{1}{c|}{\textbf{-55.44}}                         & -53.12   & -68.46 \\
\multicolumn{1}{c|}{Conference}   & -43.31 & -41.60 & \multicolumn{1}{c|}{\textbf{-45.27}}                         & -42.37 & -63.59 \\
\multicolumn{1}{c|}{SCC}          & -63.74 & -47.28 & \multicolumn{1}{c|}{\color[HTML]{FF0000}\textbf{-77.20}} & 47.54 &  54.24  \\ \midrule
\multicolumn{6}{c}{BDBR (MS-SSIM)}                                                                                              \\ \midrule
\multicolumn{1}{c|}{Surveillance} & -48.66      & -47.83 & \multicolumn{1}{c|}{\color[HTML]{FF0000}\textbf{-69.20}}                                       & -40.44      & -62.28      \\
\multicolumn{1}{c|}{Conference}   & -35.61      & -23.56 & \multicolumn{1}{c|}{\textbf{-57.07}}                                       & -33.16      & -59.55      \\
\multicolumn{1}{c|}{SCC}          & -27.30      & -4.50 & \multicolumn{1}{c|}{\color[HTML]{FF0000}\textbf{-82.41}}                                       & 12.86      & 18.18      \\ \bottomrule
\end{tabular}
\label{bdbr_classe_scc}
\end{table*}

\begin{figure*}[!t]
    \centering
    \includegraphics[width=\linewidth]{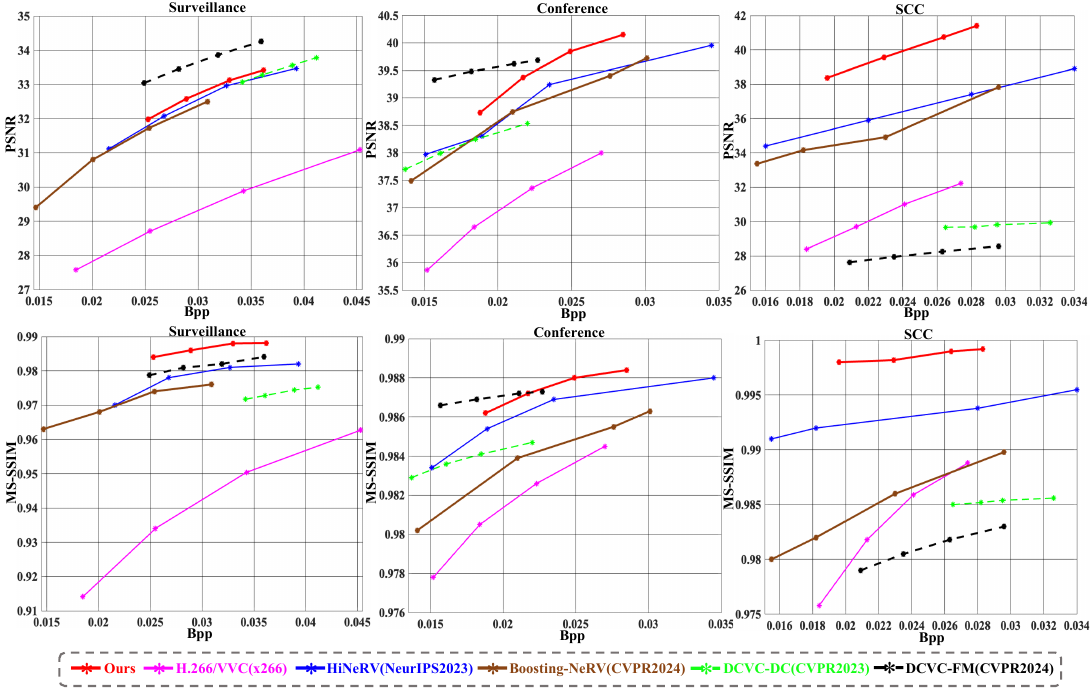}
    \caption{RD curves of our proposed method and other methods on surveillance, conference and SCC videos in terms of PSNR and MS-SSIM.}
    \vspace{-0.5cm}
    \label{rd_classe_scc}
\end{figure*}

\begin{figure*}[!t]
    \centering
    \includegraphics[width=0.98\linewidth]{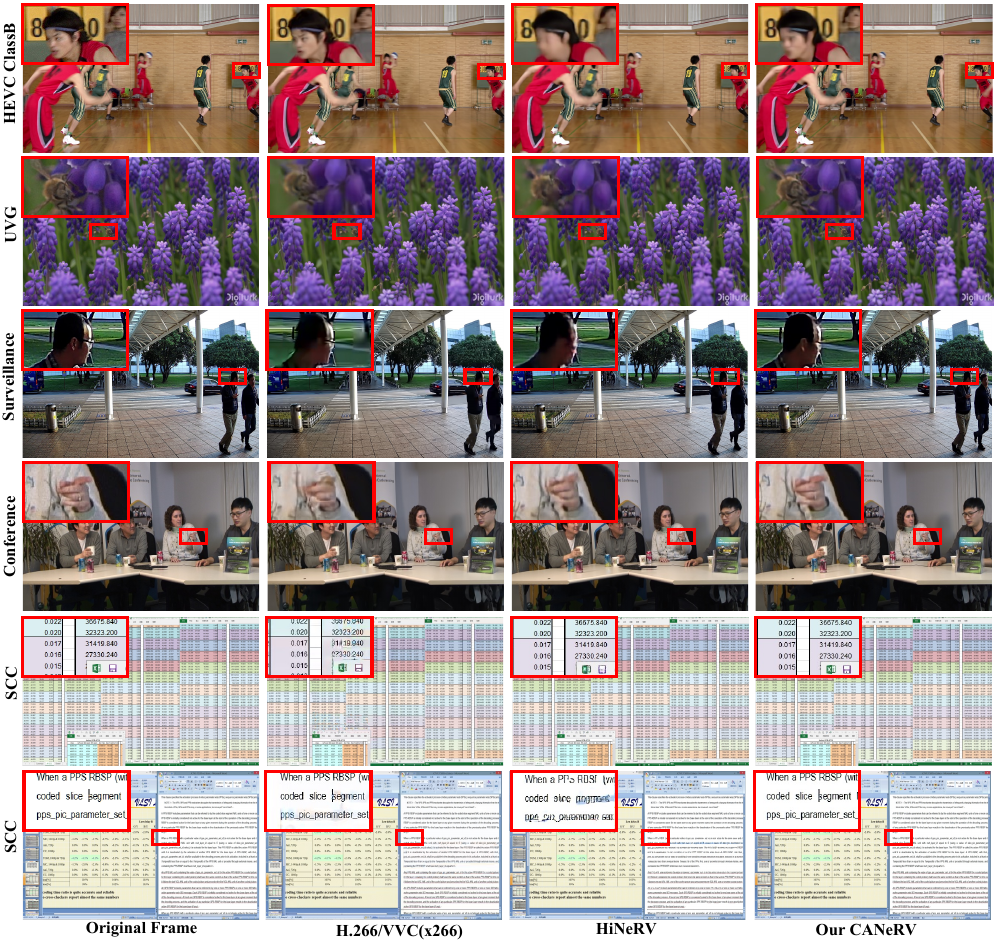}
    \caption{The visual comparison results of our proposed CANeRV with other methods. Across various sequences, our proposed CANeRV consistently demonstrates superior subjective results.}
    \label{ours_vis}
    \vspace{-0.5cm}
\end{figure*}

\subsection{Comparison Methods}
We compare our method against traditional codecs, INR-based methods and deep-learning based approaches. The state-of-the-art video codec x266 is selected as anchor method, which is an optimized implementation for video coding standard H.266/VVC~\cite{bross2021overview}. The configure of x266 is listed as follows:
\begin{itemize}
\item ffmpeg -f rawvideo -s FRAME\_RESOLUTION -i VIDEO\_NAME.yuv -c:v libvvenc -preset medium -qp QP -g FRAME\_NUM -vvenc-params IntraPeriod=10:DecodeingRefreshType=idr:Passes=1:verbosity= \\ 6:qpa=0 -f vvc SAVE\_NAME.266.
\end{itemize}

The INR-based methods include HiNeRV (NeurIPS2023)~\cite{DBLP:conf/nips/KwanGZGB23} and Boosting-NeRV (CVPR2024)~\cite{zhang2024boosting}, which are two most state-of-the-art INR-based methods in recent two years. Deep learning based methods include DCVC-DC (CVPR2023)~\cite{li2023neural} and DCVC-FM (CVPR2024)~\cite{li2024neural}. When evaluating the performance of these methods, we directly use the code and corresponding checkpoint files provided by the authors.

\begin{figure}[!t]
    \centering
    \includegraphics[width=\linewidth]{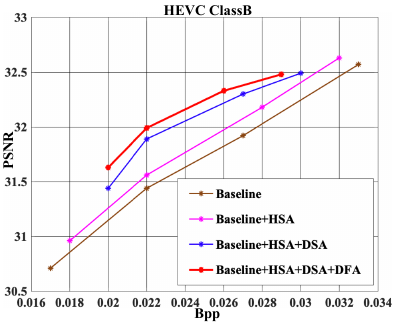}
    \caption{Ablation studies on the full architecture demonstrate improved  performance over the baseline after integrating HSA, DSA, and DFA.}
    \label{abla_arch_fig}
    \vspace{-0.5cm}
\end{figure}

\subsection{Quantitative Evaluation}
\noindent \textbf{In common scenes.} First, Tab. \ref{bdbr_classb_uvg} shows the BDBR performance for the proposed CANeRV, HiNeRV, Boosting-NeRV, DCVC-DC and DCVC-FM against H.266/VVC (x266). Notably, CANeRV surpasses the H.266/VVC (x266) standard on the HEVC ClassB and UVG datasets, marking the first time an INR-based approach has outperformed H.266/VVC (x266). This performance shows the potential of INR-based methods in video compression. Compared to the state-of-the-art INR-based method HiNeRV, CANeRV achieves around a 20\% BD rate saving, which verifies the efficiency of the proposed framework, CANeRV. Although our CANeRV framework is still inferior to the latest deep learning based methods, DCVC-DC and DCVC-FM, according to PSNR quality metric, the performance of our method is very close to them, e.g., the compression performance on UVG database. In particular, according to MS-SSIM quality metric, our proposed method has significantly outperformed DCVC-DC, achieving the best compression performance. Since DCVC-DC and DCVC-FM are trained based on large scale video database, the generalization problem will be raised when the characteristics of test videos are different from that of training videos. This problem will be explained in the following experiments on SCC videos. 

Fig. \ref{rd_classb_uvg} illustrates the RD curves for different methods to further compare their compression performance on different bitrates. Our method achieves consistent performance improvement at different bitrate scenarios compared with H.266/VVC (x266) and the state-of-the-art INR based methods. An interesting phenomenon is that the INR compression framework performs better on MS-SSIM quality metric, which is more consistent with human perceptual quality than PSNR. This is also an advantage of INR compression framework compared with traditional framework and deep learning based methods. In addition, our proposed CANeRV and other INR based video compression methods can decode any frame independently, which corresponds to random access any frame for video processing. However, traditional methods and deep learning based methods still follow intra prediction and inter prediction paradigm, which means that these methods should first
decode an intra coding frame (denoted as I-frame) before decoding the following inter prediction frames (denoted as P/B frame). Therefore, the granularity of random access is dependent on the amount of I frames.

\begin{table*}[]
\centering
\caption{Ablation studies on the entire architecture show performance improvements compared to the baseline after incorporating HSA, DSA, and DFA. Additionally, for methods like HNeRV and Boosting-NeRV, we highlight the performance gains achieved by integrating these three mechanisms, compared to their original models.}
\begin{tabular}{@{}c|ccc|c|c@{}}
\toprule
Dataset &
   Baseline+HSA &
  \begin{tabular}[c]{@{}c@{}} Baseline+HSA\\ +DSA\end{tabular} &
  \begin{tabular}[c]{@{}c@{}} Baseline+HSA\\ +DSA+DFA\end{tabular} &
  \begin{tabular}[c]{@{}c@{}} HNeRV+HSA\\ +DSA+DFA\end{tabular} &
  \begin{tabular}[c]{@{}c@{}} Boosting-NeRV+HSA\\ +DSA+DFA\end{tabular} \\ \midrule
HEVC ClassB &
  -4.78 &
  -13.46 &
  \textbf{-18.40} & -38.14
   &-21.21
   \\ \bottomrule
\end{tabular}
\label{abla_arch_table}
\vspace{-0.5cm}
\end{table*}

\noindent \textbf{In specific scenes.} Tab. \ref{bdbr_classe_scc} displays the BDBR performance for our proposed CANeRV, alongside HiNeRV, Boosting-NeRV, DCVC-DC, and DCVC-FM, against the H.266/VVC (x266) standard. Notably, CANeRV outperforms the H.266/VVC (x266) across all three datasets, showing the potential of INR-based methods in video compression. Relative to the state-of-the-art INR-based method HiNeRV, CANeRV achieves an approximate 10\% BD rate saving. Additionally, we observe that in specific scenes, our method consistently outperforms DCVC-DC. This advantage is pronounced in surveillance or conference scenes, in contrast to more dynamic scenes such as Basketball sequence in ClassB, the background exhibits minimal changes and camera transitions are infrequent, thereby simplifying the modeling process for INR networks. A similar trend is evident when using the MS-SSIM quality metric. Fig. \ref{rd_classe_scc} illustrates the RD curves for various methods, further comparing their compression performance across different bitrates. Our method consistently shows performance improvements in various bitrate scenarios compared to H.266/VVC (x266) and other advanced INR-based methods. These results demonstrate the substantial potential of INR-based video compression methods in scenarios with relatively static backgrounds. In SCC sequences, we observe that our CANeRV significantly outperforms both DCVC-DC and DCVC-FM. Our analysis suggests that the primary reason for this phenomenon is that DCVC-DC and DCVC-FM are trained using large-scale common video datasets, which may not include SCC videos. This reveals a potential generalization issue of these deep learning based methods that they cannot effectively compress videos with different characteristics from those in training sets.

\subsection{Subjective Evaluation}
As shown in Fig. \ref{ours_vis}, in the comparative analysis of visualization results, CANeRV demonstrates remarkable performance across a variety of scenarios, showing its robust adaptability and efficiency. For instance, in dynamic scenes such as those found in HEVC ClassB, our method distinctly outperforms HiNeRV by reconstructing higher-quality video frames, e.g., the much clearer face. This improvement can be attributable to our proposed DFA and HSA mechanism, which effectively captures differential information among video frames and better represents the edge structural information. In specific scenarios, such as SCC sequences where the content includes text, icons, and graphical interfaces, CANeRV excels by preserving the sharpness and readability of such elements, ensuring that the compressed video remains practical and functional for end-users. In contrast, both H.266/VVC (x266) and HiNeRV introduce significant distortions. This demonstrates the effectiveness of our CANeRV in capturing detailed structural information. Overall, our method achieves favorable subjective results across various scenes, which further shows the flexibility of CANeRV, enabled by our DSA mechanism, in finding the most suitable network structure for modeling given sequences.

\subsection{Ablation Studies}
Herein, we conduct ablation experiments on the widely used HEVC ClassB dataset to validate the effectiveness of our proposed CANeRV, which primarily contains HSA, DSA and DFA.

\subsubsection{Ablation Studies of the Whole Architecture}
We systematically validate the effectiveness of the HSA, DSA, and DFA modules integrated within the CANeRV. Each of these modules serves a distinct purpose and enhance the overall video compression performance of our INR-based video compression approach. As mentioned previously, the core focus of our paper is on proposing an adaptive mechanism to enhance the representational capabilities of INR networks. Consequently, we have not extensively explored various combinations of HSA, DSA, and DFA. Instead, we opt for a sequential integration of HSA, DSA, and DFA into the baseline INR network. The experimental results are shown in Tab. \ref{abla_arch_table} and Fig. \ref{abla_arch_fig}. Note that, the INR block design for the baseline model is same as that in the work HiNeRV.

\begin{table*}[]
\centering
\caption{Ablation Studies of our proposed HSA. We demonstrate the performance changes in BDBR (PSNR) compared to the baseline model after incorporating HSA in various ways. The baseline model in this table is consistent with the model in Tab. \ref{abla_arch_table}.}
\begin{tabular}{@{}c|cc|cccc@{}}
\toprule
Dataset &
  \begin{tabular}[c]{@{}c@{}}Baseline+HSA\\ (First-order)\end{tabular} &
  \begin{tabular}[c]{@{}c@{}}Baseline+HSA\\ (First-order+Second-order)\end{tabular} &
  \begin{tabular}[c]{@{}c@{}}Baseline+HSA\\ (Layer4)\end{tabular} &
  \begin{tabular}[c]{@{}c@{}}Baseline+HSA\\ (Layer3$\sim$4)\end{tabular} &
  \begin{tabular}[c]{@{}c@{}}Baseline+HSA\\ (Layer2$\sim$4)\end{tabular} &
  \begin{tabular}[c]{@{}c@{}}Baseline+HSA\\ (Layer1$\sim$4)\end{tabular} \\ \midrule
HEVC ClassB &
  -1.98 &
  \textbf{-4.78} &
  -4.62 &
  \textbf{-4.78} &
  0.23 &
  1.25 \\ \bottomrule
\end{tabular}
\label{abla_hsa_bdbr}
\end{table*}

\begin{table*}[]
\centering
\caption{Ablation Studies of the DFA. This table shows the performance changes when DFA is incorporated at different INR Layers.}
\begin{tabular}{@{}c|c|cccc@{}}
\toprule
Dataset &
  Baseline+HSA+DSA &
  \begin{tabular}[c]{@{}c@{}}Baseline+HSA+DSA\\ +DFA(Layer4)\end{tabular} &
  \begin{tabular}[c]{@{}c@{}}Baseline+HSA+DSA\\ +DFA(Layer3$\sim$4)\end{tabular} &
  \begin{tabular}[c]{@{}c@{}}Baseline+HSA+DSA\\ +DFA(Layer2$\sim$4)\end{tabular} &
  \begin{tabular}[c]{@{}c@{}}Baseline+HSA+DSA\\ +DFA(Layer1$\sim$4)\end{tabular} \\ \midrule
HEVC ClassB &
  -13.46 &
  \textbf{-18.40} &
  -14.32 &
  -10.23 &
  -8.23 \\ \bottomrule
\end{tabular}
\label{cft_layer}
\end{table*}

\begin{table*}[]
\centering
\caption{Ablation Studies of the DFA. This table shows the performance changes when setting different low-rank ($R$) parameters in DFA. }
\begin{tabular}{@{}cc|cccc@{}}
\toprule
Dataset &
  Baseline+HSA+DSA &
  \begin{tabular}[c]{@{}c@{}}Baseline+HSA+DSA\\ +DFA($R$=1)\end{tabular} &
  \begin{tabular}[c]{@{}c@{}}Baseline+HSA+DSA\\ +DFA($R$=3)\end{tabular} &
  \begin{tabular}[c]{@{}c@{}}Baseline+HSA+DSA\\ +DFA($R$=5)\end{tabular} &
  \begin{tabular}[c]{@{}c@{}}Baseline+HSA+DSA\\ +DFA($R$=7)\end{tabular} \\ \midrule
HEVC ClassB &
  -13.46 &
  -14.61 &
  \textbf{-18.40} &
  -13.42 &
  -9.34 \\ \bottomrule
\end{tabular}
\label{cft_R}
\vspace{-0.5cm}
\end{table*}

We begin by incorporating the HSA module into our baseline INR network. HSA is designed to assist the network in capturing fine-grained details by enhancing the structural information within video frames, thereby resulting in higher-quality video frames. As demonstrated in Tab. \ref{abla_arch_table} and Fig. \ref{abla_arch_fig}, the integration of HSA results in approximately  5\% BD rate savings since HSA only introduces a small amount of parameters for a $3 \time 3$ convolution operation, the BD rate saving mainly comes from the reconstruction quality improvement, about 0.2dB as illustrated in Fig. \ref{abla_arch_fig}. Following the integration of HSA, we add the DSA module to the INR network. DSA enables the network to adaptively modify its architecture based on the content characteristics of the video, thus further enhancing the network’s RD performance. As shown in Tab. \ref{abla_arch_table} and Fig. \ref{abla_arch_fig}, incorporating DSA results in additional 9\% BD rate savings against that of HSA, which corresponds approximately 0.4 dB improvement in PSNR as shown in Fig. \ref{abla_arch_fig}. Finally, we integrate the DFA module, which further refines the network’s capability by adapting its structure at the frame level. This module results in additional 5\% BD rate savings against the combination of HSA and DSA, which corresponds to approximately a 0.2 dB improvement in PSNR. We can see that although DFA introduces frame-level parameters, it really can well improve the reconstruction quality. In addition, the performance improvement also prove that our design for DFA is effective, which well controls the amount of parameters for the individual frame by utilizing low-rank matrix factorization.

To explore the robustness and versatility of our enhancements, we also apply these modules to two other typical INR-based video compression methods, HNeRV~\cite{DBLP:conf/cvpr/ChenGLS23a} and Boosting-NeRV~\cite{zhang2024boosting}. As shown in Tab. \ref{abla_arch_table}, the inclusion of HSA, DSA, and DFA also results in significant performance improvements. Through this series of experiments, the effectiveness of our proposed adaptive mechanism has been thoroughly validated. In subsequent ablation studies, we further verify the settings of some hyperparameters within HSA and DFA.

\subsubsection{Ablation Studies of the HSA}
In our study on HSA, we initially focus on capturing first-order structural information (such as edges) within the INR network. We then expand this approach to include both first-order and second-order structural details (which consider the curvature and continuity of edges) in the network’s processing. The ablation experiments presented in Tab. \ref{abla_hsa_bdbr} demonstrate that only the introduction of first-order structural information can achieve approximately 2\% BD rate savings. Furthermore, incorporating second-order structural information results in additional 3\% BD rate savings.

Initially, HSA is integrated only at the last layer of the CANeRV network (Layer4). Motivated by the positive results, we extend the implementation to front layers. In particular, the first to last layers (Layer1$\sim$4), the second to last (Layer2$\sim$4) and third to last (Layer3$\sim$4). The performance for these configurations is documented in the accompanying Tab. \ref{abla_hsa_bdbr}. The results indicate that incorporating HSA in the Layer4 and Layer3$\sim$4 shows similar performance. However, extending HSA to the Layer2$\sim$4 and Layer1$\sim$4 results in a noticeable decline in performance. This performance drop may be due to the lower resolution at these deeper layers, which likely impedes the network’s ability to capture detailed structural information effectively. Therefore, the additional network parameters introduced at these layers cannot enhance the reconstruction quality. This suggests that the placement of HSA within the network is crucial and the optimal benefits can be obtained when applying HSA closer to the output layer where the resolution is higher, allowing for more precise detail capture.

\subsubsection{Ablation Studies of the DFA}
In our investigation of DFA, several critical design elements are assessed: the effective layers for DFA integration and the optimal configuration of low-rank parameters. Our initial experiments focus on determining the most beneficial layers for integrating the DFA. As shown in Tab. \ref{cft_layer}, by inserting learnable parameters $\Delta W_l^t$  at various layers from the first to the last (i.e., the fourth layer), we find that the RD performance is superior when $\Delta W_l^t$ is added to the last layer. This finding shows the importance of adapting the INR network’s last layers to enhance frame-specific adaptability. Additionally, we explore the impact of different low-rank settings ($R=1,3,5,7$ in Eqn. \ref{low_Rank_Eqn}) on the performance. Results in Tab. \ref{cft_R} indicate that a rank setting of $R=3$ achieves the best performance, illustrating the efficiency of using a minimal number of parameters to capture the unique characteristics of each frame effectively.

\section{Future Work}
The promising results achieved by CANeRV in the current studies open up several interesting possibilities for further research and development. First, considering the effective integration of the HSA, DSA, and DFA modules, future work could explore deeper into hybrid integration strategies. 

In this paper, our experiments indicate that in relatively static scenes, such as surveillance or conference, CANeRV demonstrates superior performance. Therefore, further optimizing the performance of INR-based video compression methods for these specific scenarios, or identifying the most suitable video sequences for INR network modeling, may be a worthwhile direction for exploration. Such investigations could make INR-based video compression achieves more success in specific domain.

Finally, the integration of reinforcement learning or other decision-making algorithms to dynamically select the optimal configuration of network layers during encoding or decoding stage could further enhance the adaptability and performance of our proposed CANeRV. However, how to balance the compression performance and computation complexity is still a challenge for INR-based video compression.

We hope that the future research directions outlined above will motivate researchers to further refine CANeRV or contribute to the advancement of the field of INR-based video compression.

\section{Conclusion}

In this paper, we proposed an innovative Content Adaptive Neural Representation for Video Compression, named CANeRV, by introducing adaptive mechanisms to optimize network architecture and improve the represent capability of network parameters. We designed three adaptive mechanisms from three aspects including video sequence level, frame level and structure level within frames. Herein, the proposed DSA and DFA aimed to make the network adapt to variations among video sequences and video frames respectively by designing effective network parameters allocation strategies, which corresponds to the network architecture adjustment. The proposed HSA aimed to improve the network representation ability to structural information within video frames by introducing first-order and second-order structural information to supervise the learning process. Extensive experimental results and analyses have been provided in this paper, and verified the effectiveness of the proposed CANeRV.  In particular, our proposed method outperformed existing methods including the latest video coding standard H.266/VVC across diverse and extensive datasets. In addition, we also found that the INR based video compression framework is more suitable for some specific video contents, \textit{e.g.}, surveillance video, screen content video and conference video. {It is worth noting that, there is no generalization problem for INR based video compression, but deep learning based methods exposed serious generalization problem.} We hope the proposed methods and these interesting analyses can provide more insights for researchers.

{\small
\bibliographystyle{IEEEtran}
\bibliography{IEEEfull}
}

\vspace{-40pt}

\begin{IEEEbiography}
[{\includegraphics[width=1in,height=1.25in,clip,keepaspectratio]{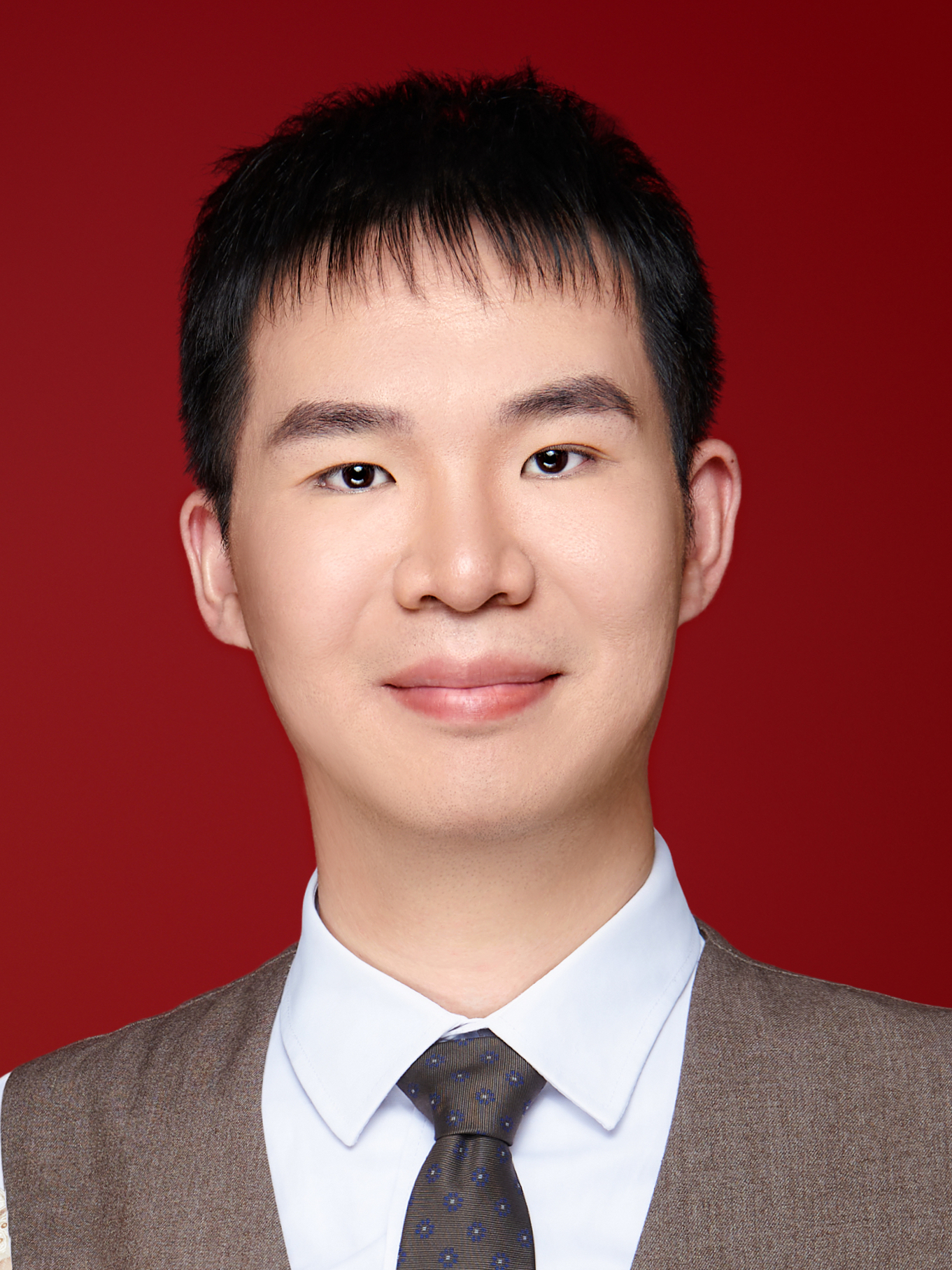}}] 
{Lv Tang} (Student Member, IEEE) received the BSc degree from the School of Information Science and Technology, Southwest Jiaotong University, China, in 2018. He received the Master's degree from the Department of Computer Science, Nanjing University, China, in 2021.
He is now pursuing a doctoral degree in School of Computer Science and Technology, University of Chinese Academy of Sciences. His research interests include computer vision, pattern recognition and video compression.
\end{IEEEbiography}

\vspace{-40pt}

\begin{IEEEbiography}
[{\includegraphics[width=1in,height=1.25in,clip,keepaspectratio]{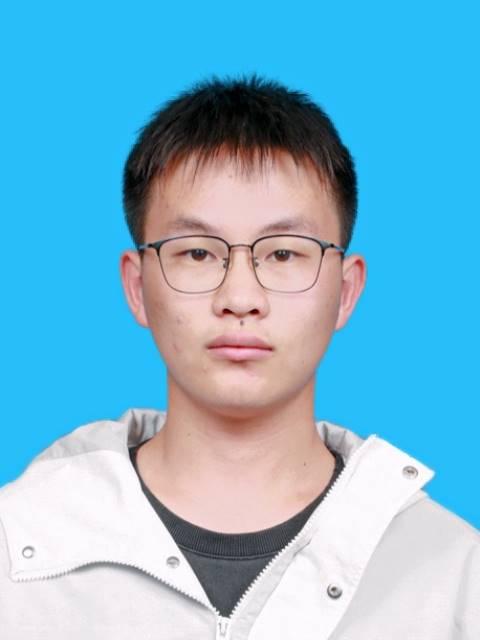}}] 
{Jun Zhu} (Student Member, IEEE) received the BSc degree from School of Software, Tsinghua University, China, in 2023.
He is now pursuing a doctoral degree in School of Computer Science and Technology, University of Chinese Academy of Sciences. His research interests include computer vision, pattern recognition and video compression.
\end{IEEEbiography}

\vspace{-40pt}
\begin{IEEEbiography}[{\includegraphics[width=1in,height=1.25in,clip,keepaspectratio]{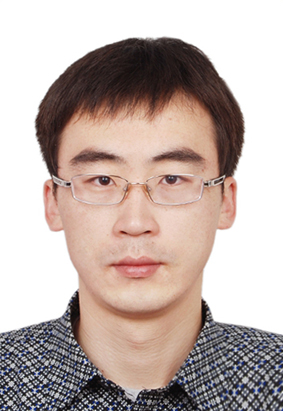}}]{Xinfeng Zhang} (Senior Member, IEEE) received the B.S. degree in computer science from the Hebei University of Technology, Tianjin, China, in 2007, and the Ph.D. degree in computer science from the Institute of Computing Technology, Chinese Academy of Sciences, Beijing, China, in 2014. From 2014 to 2017, he was a Research Fellow with the Rapid-Rich Object Search Lab, Nanyang Technological University, Singapore. From Oct. 2017 to Oct. 2018, he was a Post-Doctoral Fellow with  the School of Electrical Engineering System, University of Southern California, Los Angeles, CA, USA. From Dec. 2018 to Aug. 2019, he was a Research Fellow with the department of Computer Science, City University of Hong Kong. \par
He currently is an Assistant Professor with the School of Computer Science and Technology, University of Chinese Academy of Sciences. He authored more than 170 refereed journal/conference papers and received the Best Paper Award of IEEE Multimedia 2018, the Best Paper Award at the 2017 Pacific-Rim Conference on Multimedia (PCM) and the Best Student Paper Award in IEEE International Conference on Image Processing 2018. His research interests include video compression and processing, image/video quality assessment, and 3D point cloud processing. He
serves as an Associate Editor for the IEEE Transactions on Image Processing, Circuits and Systems for Video Technology and APSIPA Transactions on Signal and Information Processing.

\end{IEEEbiography}

\vspace{-40pt}
\begin{IEEEbiography}[{\includegraphics[width=1in,height=1.25in,clip,keepaspectratio]{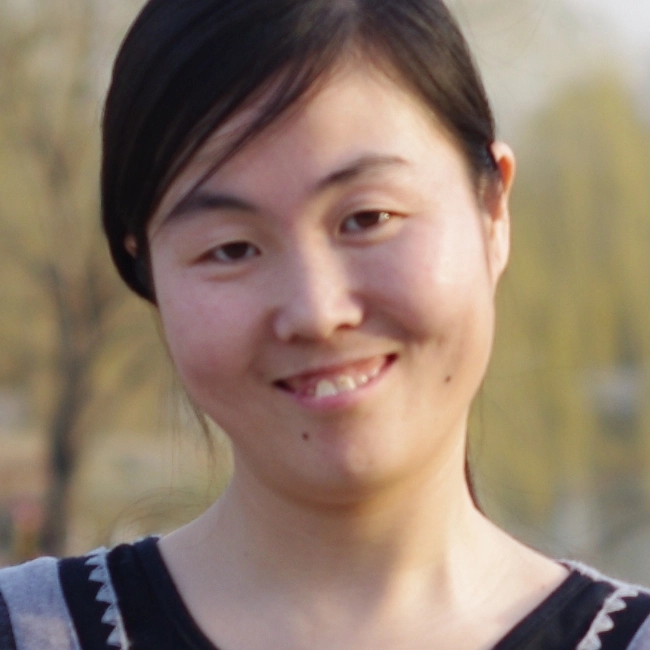}}]{Li Zhang} (Senior Member, IEEE) received the
Ph.D. degree in computer science from the Institute
of Computing Technology, Chinese Academy of
Sciences, Beijing, China, in 2009. \par
From 2009 to 2011, she held a post-doctoral
position with the Institute of Digital Media, Peking
University, Beijing. From 2011 to 2018, she was a
Senior Staff Engineer with the Multimedia Research
and Development and Standards Group, Qualcomm
Inc., San Diego, CA, USA. She is currently the
Lead of the Multimedia Laboratory, Bytedance Inc.,
San Diego. Her research interests include 2D/3D image/video coding, video
processing, and transmission. She was a Software Coordinator for Audio and
Video Coding Standard (AVS) and the 3D Extensions of High Efficiency
Video Coding (HEVC). She has authored more than 500 standardization
contributions, more than 500 granted U.S. patents, more than 100 technical
articles in related book chapters, journals, and proceedings in image/video
coding and video processing with more than 12,000 citations from Google
Scholar and best paper awards. She has been an active contributor to the
Versatile Video Coding, Advanced AVS, the IEEE 1857, 3D Video (3DV)
coding extensions of H.264/AVC and HEVC, and HEVC screen content
coding extensions. During the development of those video coding standards,
she co-chaired several ad hoc groups and core experiments. She has been
appointed as an Editor of AVS and the Main Editor of the Software Test Model
for 3DV Standards. She organized/co-chaired multiple special sessions and
grand challenges at various conferences/journals. She serves as an Associate
Editor for IEEE TRANSACTIONS ON CIRCUITS AND SYSTEMS FOR VIDEO
TECHNOLOGY and the Publicity Subcommittee Chair of the Technical Committee Member of Visual Signal Processing and Communications in IEEE
CAS Society (VSPC TC).
\end{IEEEbiography}

\vspace{-30pt}
\begin{IEEEbiography}[{\includegraphics[width=1in,height=1.25in,clip,keepaspectratio]{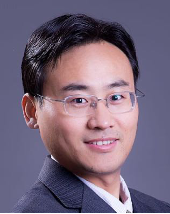}}]{Siwei Ma}
(Fellow, IEEE) received the B.S. degree from Shandong Normal University, Jinan, China, in 1999, and the Ph.D. degree in computer science from the Institute of Computing Technology, Chinese Academy of Sciences, Beijing, China, in 2005. He held a postdoctoral position with the University of Southern California, Los Angeles, CA, USA, from 2005 to 2007. He joined the School of Electronics Engineering and Computer Science, Institute of Digital Media, Peking University, Beijing, where he is currently a Professor. He has authored over 300 technical articles in refereed journals and proceedings in image and video coding, video processing, video streaming, and transmission. He served/serves as an Associate Editor for the IEEE Transactions on Circuits and Systems for Video Technology and the Journal of Visual Communication and Image Representation.
\end{IEEEbiography}

\vspace{-30pt}
\begin{IEEEbiography}[{\includegraphics[width=1in,height=1.25in,clip,keepaspectratio]{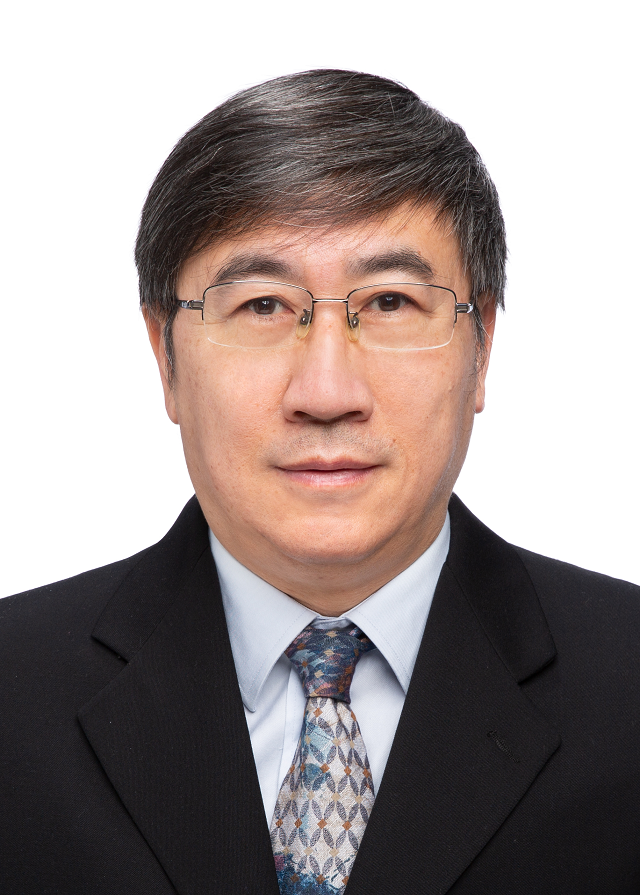}}]{Qingming Huang} (Fellow, IEEE)  received the bachelor’s degree in computer science and Ph.D. degree in computer engineering from the Harbin Institute of Technology, Harbin, China, in 1988 and 1994, respectively. He is a Professor with the University of Chinese Academy of Sciences and an Adjunct Research Professor with the Institute of Computing Technology, Chinese Academy of Sciences, Beijing, China. \par
He has authored or coauthored more than 400 academic papers in prestigious international journals,
including IEEE TRANSACTIONS ON IMAGE PROCESSING,IEEE TRANSACTIONS MULTIMEDIA, and IEEE TRANSACTIONS ON CIRCUITS AND SYSTEMS FOR VIDEO TECHNOLOGY, etc, and top-level conferences, such as ACM Multimedia, ICCV, CVPR, IJCAI, VLDB, etc. His research interests include multimedia video analysis, image processing, computer vision, and pattern recognition. He is an Associate Editor of IEEE TRANSACTIONS ON CIRCUITS AND SYSTEMS FOR VIDEO TECHNOLOGY, and Acta Automatica Sinica, and the Reviewer of various international journals, including IEEE TRANSACTIONS MULTIMEDIA, IEEE TRANSACTIONS ON CIRCUITS AND SYSTEMS FOR VIDEO TECHNOLOGY, and IEEE TRANSACTIONS ON IMAGE PROCESSING. He is a Fellow of IEEE and has served as the General Chair, Program Chair, Track Chair and TPC Member for various conferences, including ACM Multimedia, CVPR, ICCV, ICME, PCM, PSIVT, etc.
\end{IEEEbiography}

\end{document}